\documentclass[10pt, a4paper]{article}
\usepackage{lrec}
\usepackage{graphicx}
\usepackage{tabularx}
\usepackage{soul}

\usepackage{epstopdf}
\usepackage[latin1]{inputenc}

\usepackage{hyperref}
\usepackage{xstring}

\usepackage{mathtools}
\usepackage{url}
\usepackage{color}
\usepackage{lipsum}
\usepackage{enumitem}
\usepackage{microtype}

\let\oldhref\href
\renewcommand{\href}[2]{\oldhref{#1}{\hbox{#2}}}

\providecommand{\shortcite}[1]{\cite{#1}}
\newcommand{\eat}[1]{} 

\def\INCLUDEAPPENDIX{}

\title{Multi-class Hierarchical Question Classification for Multiple Choice Science Exams}

\name{Dongfang Xu$^{\ast}$, Peter Jansen$^{\ast}$, Jaycie Martin$^{\dagger}$, Zhengnan Xie$^{\dagger}$, Vikas Yadav$^{\ast}$, \\{\bf \large Harish Tayyar Madabushi$^{\ddagger}$, Oyvind Tafjord$^{\mathsection}$ and Peter Clark$^{\mathsection}$} }

\address{$^{\ast}$School of Information, University of Arizona, Tucson, AZ, USA \\
         $^{\dagger}$Department of Linguistics, University of Arizona, Tucson, AZ, USA \\
         $^{\ddagger}$School of Computer Science, University of Birmingham, Birmingham, UK \\
         $^{\mathsection}$Allen Insitute for Artificial Intelligence, Seattle, WA, USA \\
         pajansen@email.arizona.edu\\}

\abstract{
Prior work has demonstrated that question classification (QC), recognizing the \textit{problem domain}
of a question, can help answer it
more accurately. However, developing strong QC algorithms
has been hindered by the limited size and complexity of annotated data
available. To address this, we present the largest challenge dataset
for QC, containing 7,787 science exam questions paired with
detailed classification labels from a fine-grained hierarchical taxonomy of 406 problem domains.
We then show that a BERT-based model trained on
this dataset achieves a large (+0.12 MAP) gain 
compared with previous methods, while also
achieving state-of-the-art performance on benchmark
open-domain and biomedical QC datasets.
Finally, we show that using this model's predictions
of question topic significantly improves the accuracy of a
question answering system by +1.7\% P@1, with substantial
future gains possible as QC performance improves. \\ \newline \Keywords{question answering, question classification} }

\begin{document}

\maketitleabstract

\section{Introduction}

\noindent Understanding what a question is asking is one of the first steps that humans use to work towards an answer.  In the context of  question answering, question classification allows automated systems to intelligently target their inference systems to domain-specific solvers capable of addressing specific kinds of questions and problem solving methods with high confidence and answer accuracy \cite{hovy2001toward,Moldovan:2003:PIE:763693.763694}. 


\indent To date, question classification has primarily been studied in the context of open-domain TREC questions \cite{voorhees2000building}, with smaller recent datasets available in the biomedical \cite{roberts2014automatically,wasim2019multi} and education \cite{godea2018annotating} domains. The open-domain TREC question corpus is a set of 5,952 short factoid questions paired with a taxonomy developed by Li and Roth \shortcite{li2002learning} that includes 6 coarse answer types (such as \textit{entities}, \textit{locations}, and \textit{numbers}), and 50 fine-grained types (e.g. specific kinds of entities, such as \textit{animals} or \textit{vehicles}).  While a wide variety of syntactic, semantic, and other features and classification methods have been applied to this task, culminating in near-perfect classification performance \cite{madabushi2016high}, recent work has demonstrated that QC methods developed on TREC questions generally fail to transfer to datasets with more complex questions such as those in the biomedical domain \cite{roberts2014automatically}, likely due in part to the simplicity and syntactic regularity of the questions, and the ability for simpler term-frequency models to achieve near-ceiling performance \cite{xia2018novel}.

\begin{figure}[t]
	\centering
	\includegraphics[scale=0.365]{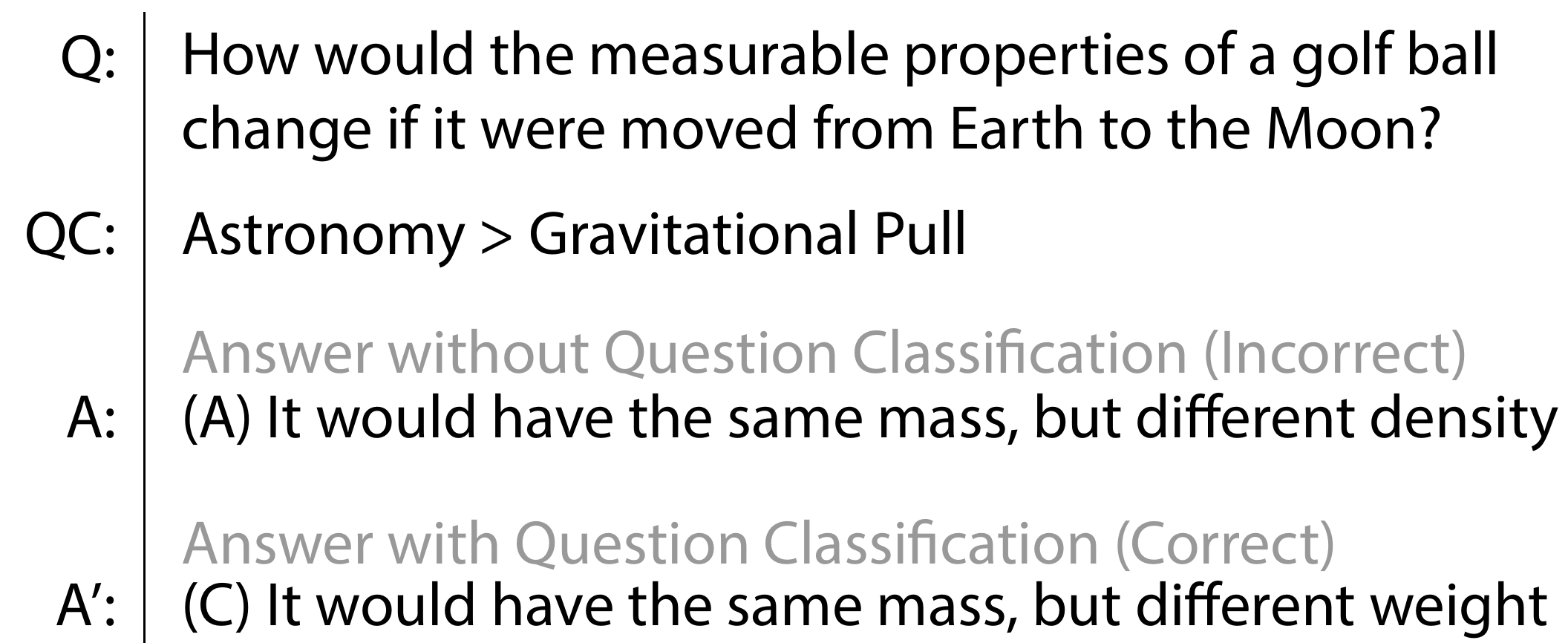}
	\caption{\small Identifying the detailed problem domain of a question (QC label) can provide an important contextual signal to guide a QA system to the correct answer (A').  Here, knowing the problem domain of \textit{Gravitational Pull} allows the model to recognize that some properties (such as weight) change when objects move between celestial bodies, while others (including density) are unaffected by such a change. \label{fig:example_questions}}
\end{figure}

In this work we explore question classification in the context of multiple choice science exams.
Standardized science exams have been proposed as a challenge task for question answering \cite{clark:2015}, as most questions contain a variety of challenging inference problems \cite{clark:2013,jansen2016:COLING}, require detailed scientific and common-sense knowledge to answer and explain the reasoning behind those answers \cite{jansen2018worldtree}, and questions are often embedded in complex examples or other distractors.
Question classification taxonomies and annotation are difficult and expensive to generate, and because of the unavailability of this data, to date most models for science questions use one or a small number of generic solvers that perform little or no question decomposition \cite[e.g.]{Khot2015ExploringML,Clark2016CombiningRS,Khashabi:2016TableILP,Khot:ACL2017,jansen2017framing}. Our long-term interest is in developing methods that intelligently target their inferences to generate both correct answers and compelling human-readable explanations for the reasoning behind those answers.  The lack of targeted solving -- using the same methods for inferring answers to spatial questions about planetary motion, chemical questions about photosynthesis, and electrical questions about circuit continuity -- is a substantial barrier to increasing performance (see Figure~\ref{fig:example_questions}).

%
%



To address this need for developing methods of targetted inference, this work makes the following contributions:
\begin{enumerate}[leftmargin=0.5cm]
\item We provide a large challenge dataset of question classification labels for 7,787 standardized science exam questions labeled using a  hierarchical taxonomy of 406 detailed problem types across 6 levels of granularity.  To the best of our knowledge this is the most detailed question classification dataset constructed by nearly an order of magnitude, while also being 30\% larger than TREC, and nearly three times the size of the largest biomedical dataset. 
\item We empirically demonstrate large performance gains of +0.12 MAP (+13.5\% P@1) on science exam question classification using a BERT-based model over five previous state-of-the art methods, while improving performance on two biomedical question datasets by 4-5\%.  This is the first model to show consistent state-of-the-art performance across multiple question classification datasets. 

\item We show predicted question labels significantly improve a strong QA model by +1.7\% P@1, where ceiling performance with perfect classification can reach +10.0\% P@1.  We also show that the error distribution of question classification matters when coupled with multiple choice QA models, and that controlling for correlations between classification labels and incorrect answer candidates can increase performance.


\end{enumerate}









\section{Related work}

\noindent Question classification typically makes use of a combination of syntactic, semantic, surface, and embedding methods.  
Syntactic patterns \cite{li2006learning,silva2011symbolic,patrick2012ontology,mishra2013question} and syntactic dependencies \cite{roberts2014automatically} have been shown to improve performance, while syntactically or semantically important words are often expanding using Wordnet hypernyms 
or Unified Medical Language System categories (for the medical domain) to help mitigate sparsity \cite{huang2008question,yu2008automatically,van2016improving}. 
Keyword identification helps identify specific terms useful for classification \cite{liu2011toward,roberts2014automatically,khashabi2017learning}.
Similarly, named entity recognizers \cite{li2002learning,neves2016biomedlat} or lists of semantically related words \cite{li2002learning,van2016improving} can also be used to establish broad topics or entity categories and mitigate sparsity, as can word embeddings \cite{kim2014cnn,lei2018novel}.  Here, we empirically demonstrate many of these existing methods do not transfer to the science domain. 

The highest performing question classification systems tend to make use of customized rule-based pattern matching \cite{lally2012question,madabushi2016high}, or a combination of rule-based and machine learning approaches \cite{silva2011symbolic}, at the expense of increased model construction time. 
  A recent emphasis on learned methods has shown a large set of CNN \cite{lei2018novel} and LSTM \cite{xia2018novel} variants achieve similar accuracy on TREC question classification, with these models exhibiting at best small gains over simple term frequency models.  These recent developments echo the observations of Roberts et al. \shortcite{roberts2014automatically}, who showed that existing methods beyond term frequency models failed to generalize to medical domain questions.
Here we show that strong performance across multiple datasets is possible using a single learned model. 

Due to the cost involved in their construction, question classification datasets and classification taxonomies tend to be small, which can create methodological challenges.
Roberts et al. \shortcite{roberts2014automatically} generated the next-largest dataset from TREC, containing 2,936 consumer health questions classified into 13 question categories.  More recently, Wasim et al. \shortcite{wasim2019multi} generated a small corpus of 780 biomedical domain questions organized into 88 categories.  In the education domain, Godea et al. \shortcite{godea2018annotating} collected a set of 1,155 classroom questions and organized these into 16 categories.  To enable a detailed study of science domain question classification, here we construct a large-scale challenge dataset that exceeds the size and classification specificity of other datasets, in many cases by nearly an order of magnitude.

%
%
\begin{table}[t]
\small
\centering
\begin{tabular}{lccc}

										&	\textbf{TREC} 	&	\textbf{GARD}	&	\textbf{ARC}	\\
Measure									&	Open		 	&	Medical			&	Science			\\
\hline
\multicolumn{2}{l}{\textit{Average per question:}}	\\
\hspace{1em} Words 						&	9.1				&	10.3			&	20.5					\\
\hspace{1em} Sentences					&	1.0				&	1.0				&	1.7						\\
\hspace{1em} Clausal Dependencies		&	0.2				&	0.6				&	0.8						\\
\hspace{1em} Prep. Dependencies			&	0.9				&	1.1				&	2.7			\vspace{1.5mm}\\

Total Questions							&	5,952			&	2,936			&	7,787		\vspace{1.5mm}\\
Question Categories						&	6 or 50			&	13				&	9 to 406	\\

\end{tabular}
\caption{\small Summary statistics comparing the surface and syntactic complexity of the TREC, GARD, and ARC datasets.
ARC questions are complex, syntactically-diverse, and paired with a detailed classification scheme developed in this work. }  
\label{tab:summarystatistics}
\end{table}

\section{Questions and Classification Taxonomy}
{\flushleft\textbf{Questions:}} We make use of the 7,787 science exam questions of the Aristo Reasoning Challenge (ARC) corpus \cite{clark2018think}, which contains standardized $3^{rd}$ to $9^{th}$ grade science questions from 12 US states from the past decade.  Each question is a 4-choice multiple choice question.  Summary statistics comparing the complexity of ARC and TREC questions are shown in Table \ref{tab:summarystatistics}.

{\flushleft\textbf{Taxonomy:}} Starting with the syllabus for the NY Regents exam, we identified 9 coarse question categories \textit{(Astronomy, Earth Science, Energy, Forces, Life Science, Matter, Safety, Scientific Method, Other)}, then through a data-driven analysis of 3 exam study guides and the 3,370 training questions, expanded the taxonomy to include 462 fine-grained categories across 6 hierarchical levels of granularity.  The taxonomy is designed to allow categorizing questions into broad curriculum topics at it's coarsest level, while labels at full specificity separate questions into narrow problem domains suitable for targetted inference methods. 
Because of its size, a subset of the classification taxonomy is shown in Table~\ref{tab:taxonomy_larger}, with the full taxonomy and class definitions included in the supplementary material.


%
%
\begin{table}[!t]
\small
\centering
\begin{tabular}{lp{6.0cm}}

\textbf{Prop.} 	&	\textbf{Category} \\
\hline
\textbf{8.0\%}		&	\textbf{		Astronomy / Celestial Events	}\\
2.7\%				&	\hspace{1em} 	Planetary/Stellar Features	\\
2.0\%				&	\hspace{1em} 	Natural Cycles and Patterns	\\
0.7\%				&	\hspace{1em} 	Planetary/Stellar Distances	\\
0.4\%				&	\hspace{1em} 	Orbits	\\

\textbf{22.4\%}		&	\textbf{		Earth Science	}\\
8.4\%				&	\hspace{1em} 	Human Impacts on the Earth	\\
6.8\%				&	\hspace{1em} 	Weather	\\
4.5\%				&	\hspace{1em} 	Geology	\\
2.4\%				&	\hspace{1em} 	Outer Structure (Atmosphere/Hydrosphere)	\\
1.3\%				&	\hspace{1em} 	Inner Structure (Crust/Mantle/Core)	\\

\textbf{7.4\%}		&	\textbf{		Energy	}\\
1.6\%				&	\hspace{1em} 	Properties of Light	\\
1.5\%				&	\hspace{1em} 	Converting Energy	\\
0.9\%				&	\hspace{1em} 	Electricity	\\
0.9\%				&	\hspace{1em} 	Sound Energy	\\
0.4\%				&	\hspace{1em} 	Potential/Kinetic Energy	\\

\textbf{3.5\%}		&	\textbf{		Forces	}\\
0.8\%				&	\hspace{1em} 	Gravity	\\
0.7\%				&	\hspace{1em} 	Friction	\\
0.5\%				&	\hspace{1em} 	Speed/Velocity	\\
0.4\%				&	\hspace{1em} 	Mechanical Energy	\\
0.3\%				&	\hspace{1em} 	Newton's Laws	\\

\textbf{34.7\%}		&	\textbf{		Life Science	}\\
16.3\%				&	\hspace{1em} 	Life Functions	\\

13.6\%				&	\hspace{2em} 	Features and their Functions	\\
5.7\%				&	\hspace{3em} 	Cellular Biology	\\
5.3\%				&	\hspace{3em} 	Animal Features and Functions	\\
3.1\%				&	\hspace{3em} 	Plant Features and Functions	\\
1.2\%				&	\hspace{4em} 	Photosynthesis	\\
0.7\%				&	\hspace{4em} 	Reproduction/Pollination	\\
0.1\%				&	\hspace{5em} 	Seed Dispersal	\\
0.4\%				&	\hspace{4em} 	Leaves	\\
0.3\%				&	\hspace{4em} 	Roots	\\

1.2\%				&	\hspace{2em} 	Environmental Effects on Development	\\
0.8\%				&	\hspace{2em} 	Responses to Environment Changes \\
0.8\%				&	\hspace{2em} 	Basic Life Functions \\

6.3\%				&	\hspace{1em} 	Interdependence/Food Chains	\\
4.7\%				&	\hspace{1em} 	Reproduction	\\
3.3\%				&	\hspace{1em} 	Adaptations and the Environment	\\
1.4\%				&	\hspace{1em} 	Continuity of Life/Life Cycle	\\

\textbf{17.0\%}		&	\textbf{		Matter	}\\
5.0\%				&	\hspace{1em} 	Chemistry	\\
2.2\%				&	\hspace{1em} 	Measurement	\\
2.4\%				&	\hspace{1em} 	Changes of State	\\
2.5\%				&	\hspace{1em} 	Properties of Materials	\\
1.8\%				&	\hspace{1em} 	Physical vs Chemical Changes	\\
1.4\%				&	\hspace{1em} 	Mixtures	\\

\textbf{1.1\%}		&	\textbf{		Safety	}\\
0.7\%				&	\hspace{1em} 	Safety Procedures	\\
0.4\%				&	\hspace{1em} 	Safety Equipment	\\

\textbf{7.6\%}		&	\textbf{		Scientific Method	}\\
5.8\%				&	\hspace{1em} 	Components of Inference	\\
0.9\%				&	\hspace{1em} 	Graphing Data	\\
0.6\%				&	\hspace{1em} 	Scientific Models	\\

\textbf{3.3\%}		&	\textbf{		Other	}\\
1.6\%				&	\hspace{1em} 	History of Science	\\

\end{tabular}
\caption{\small A subset (approximately 10\%) of our question classification taxonomy for science exams, with top-level categories in bold. The full taxonomy contains 462 categories, with 406 of these having non-zero counts in the ARC corpus. \textit{``Prop.''} represents the proportion of questions in ARC belonging to a given category. One branch of the taxonomy \textit{(Life Science $\rightarrow ... \rightarrow$ Seed Dispersal)} has been expanded to full depth.  
\label{tab:taxonomy_larger}} 
\vspace{-24pt}
\end{table}

{\flushleft\textbf{Annotation:}} Because of the complexity of the questions, it is possible for one question to bridge multiple categories -- for example, a wind power generation question may span both \textit{renewable energy} and \textit{energy conversion}.  We allow up to 2 labels per question, and found that 16\% of questions required multiple labels.  Each question was independently annotated by two annotators, with the lead annotator a domain expert in standardized exams.  Annotators first independently annotated the entire question set, then questions without complete agreement were discussed until resolution.  Before resolution, interannotator agreement (Cohen's Kappa) was $\kappa$ = 0.58 at the finest level of granularity, and $\kappa$ = 0.85 when considering only the coarsest 9 categories.  This is considered moderate to strong agreement \cite{mchugh2012interrater}.  Based on the results of our error analysis (see Section~\ref{sec:qcerroranalysis}), we estimate the overall accuracy of the question classification labels after resolution to be approximately 96\%.
 While the full taxonomy contains 462 fine-grained categories derived from both standardized questions, study guides, and exam syllabi, we observed only 406 of these categories are tested in the ARC question set.

\begin{table*}[!t]
\small
\begin{center}
\begin{tabular}{llp{0.01cm}ccccccc}
	 	&	& &	\multicolumn{7}{c}{ARC Science Exams}	\\
Adapted From								&	Model								& & L1 		&	L2		&	L3		&	L4		&	L5		&	L6		& Gain (L6)	\\
\hline
											& Unigram Model 						& & 0.885	& 0.714		& 0.602		& 0.535		& 0.503		& 0.490		& \\
Li and Roth \shortcite{li2002learning} 		& Uni+Bi+POS+Hier \textit{(UBPH)}	& & 0.903	& 0.759		& 0.644		& 0.582		& 0.549		& 0.535		& Baseline\\
Van-tu et al. \shortcite{van2016improving}	& UBPH+WordNet Expansion				& & 0.901	& 0.755		& 0.645		& 0.582		& 0.552		& 0.535		& --\\
Roberts et al. \shortcite{roberts2014automatically}	& UBPH+Dependencies				& & 0.906	& 0.760		& 0.645		& 0.583		& 0.549		& 0.536		& --\\

He et al. \shortcite{he2015multi}			& MP-CNN								& & 0.908	& 0.757		& 0.654		& 0.597		& 0.563		& 	0.532	& --\\

Khashabi et al. \shortcite{khashabi2017learning}& UBPH+Essential Terms					& & 0.913	& 0.774		& 0.666		& 0.607		& 0.575		& 0.564		& +0.03$^{*}$\\
This Work							 		& BERT-QC 								& & \textbf{0.942}	& \textbf{0.841}		& \textbf{0.745}		& \textbf{0.684}		& \textbf{0.664}		& \textbf{0.654}		& \textbf{+0.12$^{*}$}\\
%
~\\
\multicolumn{2}{l}{\textit{Number of Categories}}						& &	\textit{9}	&	\textit{88}	& \textit{243} &	\textit{335}	&	\textit{379}	&	\textit{406}		\\
 
\end{tabular}
\caption{\small Results of the empirical evaluation on each of the question classification models on the ARC science question dataset, broken down by classification granularity (\textit{coarse (L1)} to \textit{fine (L6)}).  Performance reflects mean average precision (MAP), where a duplicate table showing P@1 is included in the appendix.  The best model at each level of granularity is shown in bold. * signifies that a given model is significantly better than baseline performance at full granularity $(p<0.01)$. \label{tab:results}}
\end{center}

\end{table*}

\section{Question Classification Models}

\subsection{Question Classification on Science Exams}

\noindent We identified 5 common models in previous work primarily intended for learned classifiers rather than hand-crafted rules.
We adapt these models to a multi-label hierarchical classification task by training a series of one-vs-all binary classifiers \cite{tsoumakas2007multi}, one for each label in the taxonomy.  With the exception of the CNN  and BERT models, following previous work \cite[e.g.]{silva2011symbolic,roberts2014automatically,xia2018novel} we make use of an SVM classifier using the \texttt{LIBSvM} framework \cite{chang2011libsvm} with a linear kernel. Models are trained and evaluated from coarse to fine levels of taxonomic specificity. At each level of taxonomic evaluation, a set of non-overlapping confidence scores for each binary classifier are generated and sorted to produce a list of ranked label predictions. We evaluate these ranks using Mean Average Precision \cite[see]{manning08}. 
ARC questions are evaluated using the standard 3,370 questions for training, 869 for development, and 3,548 for testing.



{\flushleft\textbf{N-grams, POS, Hierarchical features:}} A baseline bag-of-words model incorporating both tagged and untagged unigrams and bigams.  We also implement the hierarchical classification feature of Li and Roth \cite{li2002learning}, where for a given question, the output of the classifier at coarser levels of granularity serves as input to the classifier at the current level of granularity.  

{\flushleft\textbf{Dependencies:}} Bigrams of Stanford dependencies \cite{de2008stanford}.  For each word, we create one unlabeled bigram for each outgoing link from that word to it's dependency \cite{patrick2012ontology,roberts2014automatically}.

{\flushleft\textbf{Question Expansion with Hypernyms:}} We perform hypernym expansion \cite{huang2008question,silva2011symbolic,roberts2014automatically} by including WordNet hypernyms \cite{fellbaum1998wordnet} for the root dependency word, and words on it's direct outgoing links.  WordNet sense is identified using Lesk word-sense disambiguation \cite{lesk1986automatic}, using question text for context.  We implement the heuristic of Van-tu et al. \shortcite{van2016improving}, where more distant hypernyms receive less weight.

{\flushleft\textbf{Essential Terms:}} Though not previously reported for QC, we make use of unigrams of keywords extracted using the Science Exam Essential Term Extractor of Khashabi et al. \shortcite{khashabi2017learning}.  For each keyword, we create one binary unigram feature.


{\flushleft\textbf{CNN:}} Kim \shortcite{kim2014cnn} demonstrated near state-of-the-art performance on a number of sentence classification tasks (including TREC question classification) by using pre-trained word embeddings \cite{mikolov2013distributed} as feature extractors in a CNN model. Lei et al. \shortcite{lei2018novel}  showed that 10 CNN variants perform within +/-2\% of Kim's \shortcite{kim2014cnn} model on TREC QC.  We report performance of our best CNN model based on the MP-CNN architecture\footnote{\url{https://github.com/castorini/Castor}} of Rao et al. \cite{rao2016noise}, which works to establish the similarity between question text and the definition text of the question classes. 
We adapt the MP-CNN model, which uses a ``Siamese'' structure \cite{he2015multi}, to create separate representations for both the question and the question class.  The model then makes use of a triple ranking loss function to minimize the distance between the representations of questions and the correct class while simultaneously maximising the distance between questions and incorrect classes. We optimize the network using the method of Tu \shortcite{tu2018experimental}.


{\flushleft\textbf{BERT-QC (This work):}} We make use of BERT \cite{devlin2018bert}, a language model using bidirectional encoder representations from transformers, in a sentence-classification configuration.  As the original settings of BERT do not support multi-label classification scenarios, and training a series of 406 binary classifiers would be computationally expensive, we use the duplication method of Tsoumakas et al. \shortcite{tsoumakas2007multi} where we enumerate multi-label questions as multiple single-label instances during training by duplicating question text, and assigning each instance one of the multiple labels.  Evaluation follows the standard procedure where we generate a list of ranked class predictions based on class probabilities, and use this to calculate Mean Average Precision (MAP) and Precision@1 (P@1).  As shown in Table~\ref{tab:results}, this BERT-QC model achieves our best question classification performance, significantly exceeding baseline performance on ARC by 0.12 MAP and 13.5\% P@1. 


\subsection{Comparison with Benchmark Datasets}
%
%
\begin{table}[!t]
\small
\centering
\begin{tabular}{lccc}

\textbf{}											&						& \multicolumn{2}{c}{\textbf{TREC}}\\
\textbf{Model}										&	\textbf{Desc.}		& \textbf{Coarse}		&	\textbf{Fine} \\
\hline
\multicolumn{2}{l}{\textit{Learned Models}}	\\
~~Li and Roth \shortcite{li2002learning}			&	SNoW				&	91.0\%				&	84.2\%			\\
~~Kim \shortcite{kim2014cnn}						&	CNN 				&	93.6\%				&	--				\\
~~Xia et al. \shortcite{xia2018novel}				&	TF-IDF				&	94.8\%				&	--				\\
~~Van-tu et al. \shortcite{van2016improving}		&	SVM					&	95.2\%				&	91.6\%			\\
~~Xia et al. \shortcite{xia2018novel}				&	LSTM				&	95.8\%				&	--				\\
~~Lei et al. \shortcite{lei2018novel}				&	RR-CNN				&	95.8\%				&	--				\\
~~This Work											&	BERT-QC				&	96.2\%				&	\textbf{92.0\%}	\\
~~Xia et al. \shortcite{xia2018novel}\footnote{text}&	Att-LSTM			&	\textbf{98.0\%}		&	--				\\
~\\
\multicolumn{2}{l}{\textit{Rule Based Models}}	\\
~~da Silva et al. \shortcite{silva2011symbolic}		&	Rules				&	95.0\%				&	90.8\%			\\
~~Madabushi et al. \shortcite{madabushi2016high}	&	Rules				&	--					&	97.2\%			\\

~\\
\multicolumn{2}{l}{\textit{Number of Categories}}							&	\textit{6}			&	\textit{50}		\\

\end{tabular}
\caption{\small Performance of BERT-QC on the TREC-6 (6 coarse categories) and TREC-50 (50 fine-grained categories) question classification task, in context with recent learned or rule-based models.  Bold values represent top reported learned model performance.  BERT-QC achieves performance similar to or exceeding the top reported non-rule-based models. \label{tab:treccomparison}}  
\end{table}
\noindent Apart from term frequency methods, question classification methods developed on one dataset generally do not exhibit strong transfer performance to other datasets \cite{roberts2014automatically}.  
While BERT-QC achieves large gains over existing methods on the ARC dataset, 
here we demonstrate that BERT-QC also matches state-of-the-art performance on TREC \cite{li2002learning}, while surpassing state-of-the-art performance on the GARD corpus of consumer health questions \cite{roberts2014automatically} and MLBioMedLAT corpus of biomedical questions \cite{wasim2019multi}.  As such, BERT-QC is the first model to achieve strong performance across more than one question classification dataset. 

\subsubsection{TREC Question Classification}

\noindent TREC question classification\footnote{\url{http://cogcomp.org/Data/QA/QC/}} is divided into separate coarse and fine-grained tasks centered around inferring the expected answer types of short open-domain factoid questions. TREC-6 includes 6 coarse question classes (\textit{abbreviation, entity, description, human, location, numeric}), while TREC-50 expands these into 50 more fine-grained types.

TREC question classification methods can be divided into those that learn the question classification task, and those that make use of either hand-crafted or semi-automated syntactic or semantic extraction rules to infer question classes.  To date, the best reported accuracy for learned methods is 98.0\% by Xia et al. \shortcite{xia2018novel} for TREC-6, and 91.6\% 
by Van-tu et al. \cite{van2016improving} for TREC-50\footnote{Model performance is occasionally reported only on TREC-6 rather than the more challenging TREC-50, making direct comparisons between some algorithms difficult.}.  Madabushi et al. \shortcite{madabushi2016high} achieve the highest to-date performance on TREC-50 at 97.2\%, using rules that leverage the strong syntactic regularities in the short TREC factoid questions. 

We compare the performance of BERT-QC with recently reported performance on this dataset in Table~\ref{tab:treccomparison}.  
BERT-QC achieves state-of-the-art performance on fine-grained classification (TREC-50) for a learned model at 92.0\% accuracy, and near state-of-the-art performance on coarse classification (TREC-6) at 96.2\% accuracy.\footnote{Xia et al. \shortcite{xia2018novel} also report QC performance on MS Marco \cite{nguyen2016ms}, a million-question dataset using 5 of the TREC-6 labels.  We believe this to be in error as MS Marco QC labels are automatically generated.  Still, for purposes of comparison, BERT-QC reaches 96.2\% accuracy, an increase of +3\% over Xia et al. \shortcite{xia2018novel}'s best model.}


%

%
%
\begin{table}[!t]
\small
\centering
\begin{tabular}{lccc}

\textbf{Model}										&	\textbf{Desc.}		&	\textbf{Accuracy} \\
\hline
\multicolumn{2}{l}{\textit{Learned Models}}	\\
~~Roberts et al. \shortcite{roberts2014automatically}	&	Bag of Words	&	76.9\%			\\
~~Roberts et al. \shortcite{roberts2014automatically}	&	CQT2/SVM		&	80.4\%			\\
~~This Work											&	BERT-QC				&	\textbf{84.9\%}\\
\end{tabular}
\caption{\small Performance of BERT-QC on the GARD consumer health question dataset, which contains 2,937 questions labeled with 13 medical question classification categories. Following Roberts et al. \shortcite{roberts2014automatically}, this dataset was evaluated using 5-fold crossvalidation. \label{tab:gardcomparison}}  
\end{table}

%
%
\begin{table}[!t]
\small
\centering
\begin{tabular}{lcccc}

\textbf{Model}										&	\textbf{Desc.}		&	\textbf{$\mu$F1} 	&	\textbf{Accuracy} \\
\hline
\multicolumn{2}{l}{\textit{Learned Models}}	\\
~~Wasim et al. \shortcite{wasim2019multi}			&	SSVM				&	0.42				&	0.37			\\
~~Wasim et al. \shortcite{wasim2019multi}			&	LPLR				&	0.47				&	0.42			\\
~~Wasim et al. \shortcite{wasim2019multi}			&	FDSF				&	0.50				&	0.45			\\
~~This Work											&	BERT-QC				&	\textbf{0.55}		&	\textbf{0.48}\\
\end{tabular}
\caption{\small Performance of BERT-QC on the MLBioMedLAT biomedical question dataset, which contains 780 questions labeled with 88 medical question classification categories. Following Wasim et al. \shortcite{wasim2019multi}, this dataset was evaluated using 10-fold crossvalidation. Micro-F1 ($\mu$F1) and Accuracy follow Wasim et al.'s definitions for multi-label tasks.\label{tab:biomedcomparison}}  
\end{table}

%

\subsubsection{Medical Question Classification}

\noindent Because of the challenges with collecting biomedical questions, the datasets and classification taxonomies tend to be small, and rule-based methods often achieve strong results \cite[e.g.]{sarrouti2015biomedical}.  Roberts et al. \shortcite{roberts2014automatically} created the largest biomedical question classification dataset to date, annotating 2,937 consumer health questions drawn from the Genetic and Rare Diseases (GARD) question database with 13 question types, such as \textit{anatomy, disease cause, diagnosis, disease management,} and \textit{prognoses}. 
Roberts et al.~\shortcite{roberts2014automatically} found these questions largely resistant to learning-based methods developed for TREC questions.  Their best model (CPT2), shown in Table~\ref{tab:gardcomparison}, makes use of stemming and lists of semantically related words and cue phrases to achieve 80.4\% accuracy.  BERT-QC reaches 84.9\% accuracy on this dataset, an increase of +4.5\% over the best previous model.  We also compare performance on the recently released MLBioMedLAT dataset \cite{wasim2019multi}, a multi-label biomedical question classification dataset with 780 questions labeled using 88 classification types drawn from 133 Unified Medical Language System (UMLS) categories.  Table~\ref{tab:biomedcomparison} shows BERT-QC exceeds their best model, focus-driven semantic features (FDSF), by +0.05 Micro-F1 and +3\% accuracy.

%
%
\begin{table*}[!t]
\small
\centering
\begin{tabular}{cl}

\textbf{Proportion}	&	\textbf{Error Type}\\
\hline
46\%	&	Question contains words correlated with incorrect class												\\
35\%	&	Predicted class is nearly correct, and distance 1 from gold class (different leaf node selected in taxonomy)	\\
25\%	&	Predicted class is highly correlated with an incorrect multiple choice answer						\\
18\%	&	Predicted class and gold class are on different aspects of similar topics/otherwise correlated		\\
10\%	&	Annotation: Gold label appears incorrect, predicted label is good.	\\
8\%		&	Annotation: Predicted label is good, but not in gold list.	\\
8\%		&	Correctly predicting the gold label may require knowing the correct answer to the question.	\\
\end{tabular}
\caption{\small BERT-QC Error Analysis: Classes of errors for 50 randomly selected questions from the development set where BERT-QC did not predict the correct label.  These errors reflect the BERT-QC model trained and evaluated with terms from both the question and all multiple choice answer candidates. Questions can occupy more than one error category, and as such proportions do not sum to 100\%. \label{tab:qcerrors1}}  
\end{table*}

\subsection{Error Analysis}
\label{sec:qcerroranalysis}

\noindent We performed an error analysis on 50 ARC questions where the BERT-QC system did not predict the correct label, with a summary of major error categories listed in Table~\ref{tab:qcerrors1}.

{\flushleft\textbf{Associative Errors:}} In 35\% of cases, predicted labels were nearly correct, differing from the correct label only by the finest-grained (leaf) element of the hierarchical label (for example, predicting \textit{Matter $\rightarrow$ Changes of State $\rightarrow$ Boiling} instead of \textit{Matter $\rightarrow$ Changes of State $\rightarrow$ Freezing}).  The bulk of the remaining errors were due to questions containing highly correlated words with a different class, or classes themselves being highly correlated.  For example, a specific question about \textit{Weather Models} discusses ``environments'' changing over ``millions of years'', where discussions of environments and long time periods tend to be associated with questions about \textit{Locations of Fossils}. Similarly, a question containing the word ``evaporation'' could be primarily focused on either \textit{Changes of State} or the \textit{Water Cycle} (cloud generation), and must rely on knowledge from the entire question text to determine the correct problem domain.  We believe these associative errors are addressable technical challenges that could ultimately lead to increased performance in subsequent models.

{\flushleft\textbf{Errors specific to the multiple-choice domain:}} 
We observed that using both question and all multiple choice answer text produced large gains in question classification performance -- for example, BERT-QC performance increases from 0.516 (question only) to 0.654 (question and all four answer candidates), an increase of 0.138 MAP.  Our error analysis observed that while this substantially increases QC performance, it changes the \textit{distribution of errors} made by the system.  Specifically, 25\% of errors become highly correlated with an incorrect answer candidate, which (we show in Section \ref{sec:qawithqc}) can reduce the performance of QA solvers.\footnote{When a model is trained using only question text (instead of both question and answer candidate text), the distribution of these highly-correlated errors changes to the following: 17\% chose the correct label, 17\% chose the same label, and 66\% chose a different label not correlated with an incorrect answer candidate.}

\section{Question Answering with QC Labels}
\label{sec:qawithqc}

\noindent Because of the challenges of errorful label predictions correlating with incorrect answers, it is difficult to determine the ultimate benefit a QA model might receive from reporting QC performance in isolation.
Coupling QA and QC systems can often be laborious -- either a large number of independent solvers targeted to specific question types must be constructed \cite[e.g.]{Minsky:1986:SM}, or an existing single model must be able to productively incorporate question classification information.  Here we demostrate the latter -- that a BERT QA model is able to incorporate question classification information through query expansion. 

BERT \cite{devlin2018bert} recently demonstrated state-of-the-art performance on benchmark question answering datasets such as SQUaD \cite{rajpurkar2016squad}, and near human-level performance on SWAG \cite{zellers2018swag}.  Similarly, Pan et al. \shortcite{pan2019improving} demonstrated that BERT achieves the highest accuracy on the most challenging subset of ARC science questions. We make use of a BERT QA model using the same QA paradigm described by Pan et al. \shortcite{pan2019improving}, where QA is modeled as a next-sentence prediction task that predicts the likelihood of a given multiple choice answer candidate following the question text.  We evaluate the question text and the text of each multiple choice answer candidate separately, where the answer candidate with the highest probablity is selected as the predicted answer for a given question.  Performance is evaluated using Precision@1 \cite{manning08}.   Additional model details and hyperparameters are included in the \textit{Appendix}. 


%
%
\begin{table}[t]
\small
\centering
\begin{tabular}{l}

\hline
\textit{Original Question Text}	\\
~~~What happens to water molecules during the \\
~~~~boiling process? \\
~\\
\textit{Expanded Text (QC Label)}	\\
~~~Matter Changes of State Boiling What happens to \\
~~~~water molecules during the boiling process? \\
\hline

\end{tabular}
\caption{\small An example of the query expansion technique for question classification labels, where the definition text for the QC label is appended to the question. Here, the gold label for this question is ``MAT\_COS\_BOILING'' \textit{(Matter $\rightarrow$ Changes of State $\rightarrow$ Boiling)}. \label{tab:expansionexample}}  
\end{table}

We incorporate QC information into the QA process by implementing a variant of a query expansion model \cite{qiu1993concept}.  Specifically, for a given \textit{\{question, QC\_label\}} pair, we expand the question text by  concatenating the definition text of the question classification label to the start of the question.  We use of the top predicted question classification label for each question.  Because QC labels are hierarchical, we append the label definition text for each level of the label $L_1 ... L_n$.  
An exampe of this process is shown in Table~\ref{tab:expansionexample}.


 
%
%
\begin{figure}[t]
	\centering
	\includegraphics[scale=0.30]{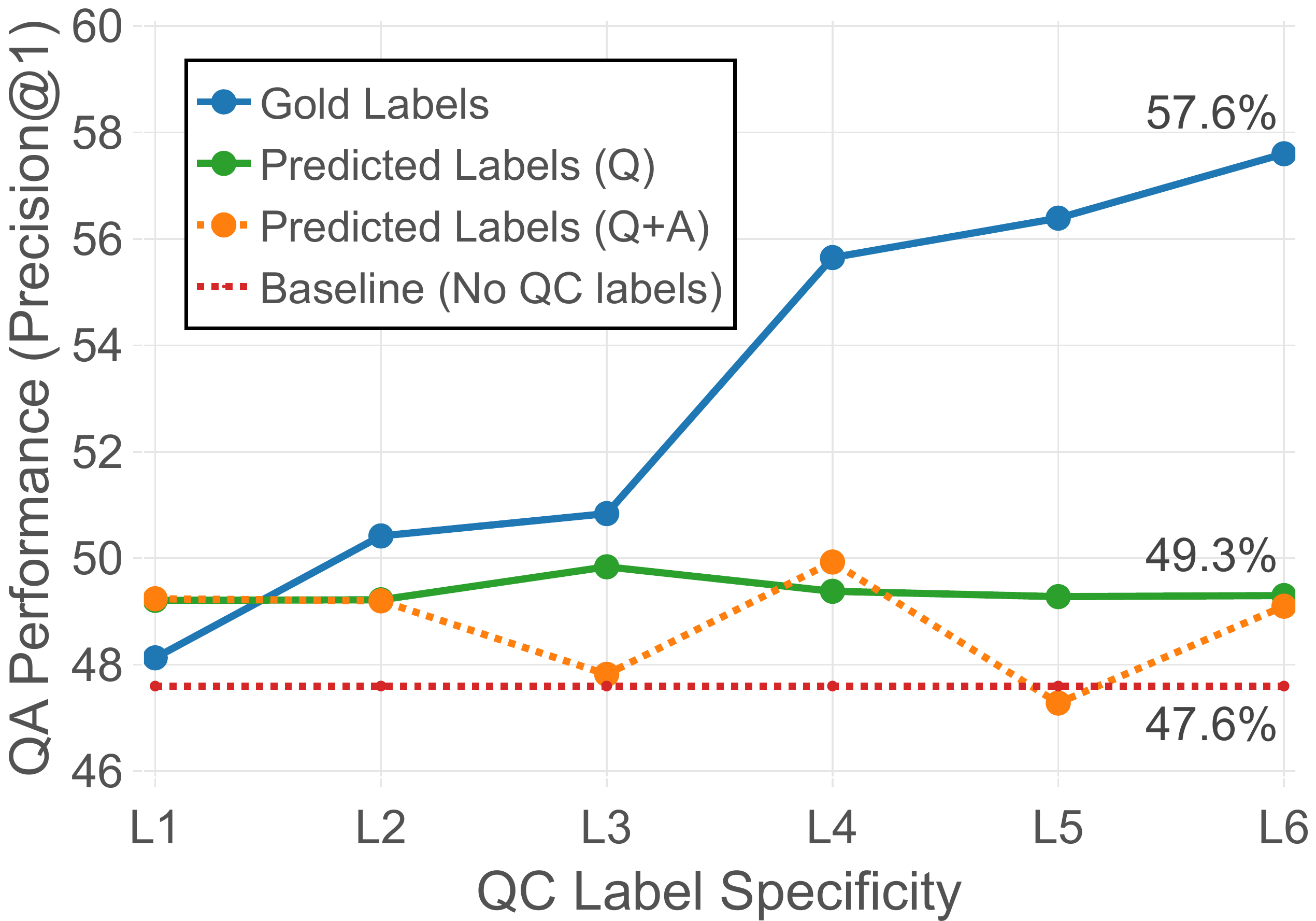}
	\caption{\small Question answering performance (the proportion of questions answered correctly) for models that include question classification labels using query expansion, compared to a no-label baseline model.  While BERT-QC trained using question and answer text achieves higher QC performance, it leads to unstable QA performance due to it's errors being highly correlated with incorrect answers.  Predicted labels using BERT-QC (question text only) show a significant increase of +1.7\% P@1 at L6 $(p<0.01)$.  Models with gold labels show the ceiling performance of this approach with perfect question classification performance.  Each point represents the average of 10 runs. \label{fig:qaperformance}}
\end{figure}

 Figure~\ref{fig:qaperformance} shows QA peformance using predicted labels from the BERT-QC model, compared to a baseline model that does not contain question classification information. As predicted by the error analysis, while a model trained with question and answer candidate text performs better at QC than a model using question text alone, a large proportion of the incorrect predictions become associated with a negative answer candidate, reducing overall QA performance, and highlighting the importance of evaluating QC and QA models together.  When using BERT-QC trained on question text alone, at the finest level of specificity (L6) where overall question classification accuracy is 57.8\% P@1, question classification significantly improves QA performance by +1.7\% P@1 $(p<0.01)$.  Using gold labels shows ceiling QA performance can reach +10.0\% P@1 over baseline, demonstrating that as question classification model performance improves, substantial future gains are possible.  An analysis of expected gains for a given level of QC performance is included in the \textit{Appendix}, showing approximately linear gains in QA performance above baseline for QC systems able to achieve over 40\% classification accuracy.  Below this level, the decreased performance from noise induced by incorrect labels surpasses gains from correct labels.





\subsection{Automating Error Analyses with QC}

%
%
\begin{table}[!t]
\small
\centering
\begin{tabular}{lcc}

\textbf{}					& \textbf{QA}		&		\textbf{} \\
\textbf{Question Category}	& \textbf{Accuracy}	&		\textbf{N} \\
\hline
\textit{Strong Performance}	\\
~~~Life - Human Health					&	73\%		&	11	\\
~~~Forces - Friction					&	71\%		&	7	\\
~~~Energy - Sound 						&	71\%		&	7	\\
~~~Energy - Light						&	70\%		&	10	\\
~~~Matter - Material Properties		&	68\%		&	22	\\
~~~Forces - Gravity					&	66\%		&	9	\\
~~~Science - Scientific Models			&	66\%		&	9	\\
~~~Earth - Inner Core					&	64\%		&	33	\\
~~~Astronomy - Natural Cycles			&	64\%		&	11	\\
~~~Energy - Device Use					&	63\%		&	8	\\
~~~Energy - Waves						&	63\%		&	8	\\
~~~Earth - Weather						&	62\%		&	74	\\
~~~Energy - Conversion					&	62\%		&	13	\\
~~~Energy - Thermal					&	60\%		&	10	\\
~\\

\textit{Above Average Performance}	\\
~~~Earth - Geology						&	58\%		&	38	\\
~~~Science - Graphs					&	56\%		&	9	\\
~~~Life - Environmental Adaptations	&	53\%		&	32	\\
~~~Matter - Phys./Chemical Changes	&	53\%		&	17	\\
~~~Astronomy - Features				&	52\%		&	27	\\
~\\

\textit{Approximately Average Performance}	\\
~~~Earth - Human Impacts				&	51\%		&	84	\\
~~~Matter - Chemistry					&	50\%		&	46	\\
~~~Life - Features and Functions		&	49\%		&	176	\\
~\\

\textit{Below Average Performance}	\\

~~~Science - Scientific Inference		&	47\%		&	58	\\
~~~Life - Food Chains					&	44\%		&	54	\\
~~~Astronomy - Celestial Distances		&	44\%		&	9	\\
~~~Life - Reproduction					&	41\%		&	41	\\
~~~Matter - Measurement				&	40\%		&	15	\\
~~~Life - Classification				&	38\%		&	13	\\
~~~Matter - Changes of State			&	29\%		&	21	\\
~\\
\textit{Below Chance Performance}	\\
~~~Earth - Outer Core					&	17\%		&	12	\\
~~~Safety - Safety Procedures			&	7\%			&	14	\\

\end{tabular}
\caption{\small Analysis of question answering performance on specific question classes on the BERT-QA model (L6). Question classes in this table are at the L2 level of specificity.  Performance is reported on the development set, where N represents the total number of questions within a given question class. \label{tab:qaqcspecificperformance} }  
\end{table}

\noindent Detailed error analyses for question answering systems are typically labor intensive, often requiring hours or days to perform manually.  As a result  error analyses are typically completed infrequently, in spite of their utility to key decisions in the algortithm or knowledge construction process.  Here we show having access to detailed question classification labels specifying fine-grained problem domains provides a mechanism to automatically generate error analyses in seconds instead of days. 


To illustrate the utility of this approach, Table~\ref{tab:qaqcspecificperformance} shows the performance of the BERT QA+QC model broken down by specific question classes.  This allows automatically identifying a given model's strengths -- for example, here questions about \textit{Human Health}, \textit{Material Properties}, and \textit{Earth's Inner Core} are well addressed by the BERT-QA model, and achieve well above the average QA performance of 49\%.  Similarly, areas of deficit include \textit{Changes of State}, \textit{Reproduction}, and \textit{Food Chain Processes} questions, which see below-average QA performance.  The lowest performing class, \textit{Safety Procedures}, demonstrates that while this model has strong performance in many areas of scientific reasoning, it is \textit{worse than chance} at answering questions about safety, and would be inappropriate to deploy for safety-critical tasks.

 While this analysis is shown at an intermediate (L2) level of specificity for space, more detailed analyses are possible.  For example, overall QA performance on \textit{Scientific Inference} questions is near average (47\%), but increasing granularity to L3 we observe that questions addressing 
 \textit{Experiment Design} or \textit{Making Inferences} -- challenging questions even for humans -- perform poorly (33\% and 20\%) when answered by the QA system.  This allows a system designer to intelligently target problem-specific knowledge resources and inference methods to address deficits in specific areas.

\section{Conclusion}

\noindent Question classification can enable targetting question answering models, but is challenging to implement with high performance without using rule-based methods.  
In this work we generate the most fine-grained challenge dataset for question classification, using complex and syntactically diverse questions, 
and show gains of up to 12\% are possible with our question classification model across datasets in open, science, and medical domains.
This model is the first demonstration of a question classification model achieving state-of-the-art results across benchmark datasets in open, science, and medical domains.  
We further demonstrate attending to question type can significantly improve question answering performance, with large gains possible as quesion classification performance improves. 
Our error analysis suggests that developing high-precision methods of question classification independent of their recall can offer the opportunity to incrementally make use of the benefits of question classification without suffering the consequences of classification errors on QA performance. 

\section{Resources}
\noindent Our Appendix and supplementary material (available at \url{http://www.cognitiveai.org/explanationbank/}) includes data, code, experiment details, and negative results.

\newpage

\section{Acknowledgements}
The authors wish to thank Elizabeth Wainwright and Stephen Marmorstein for piloting an earlier version of the question classification annotation.  We thank the Allen Insitute for Artificial Intelligence and National Science Founation (NSF 1815948 to PJ) for funding this work.

\ifdefined\INCLUDEAPPENDIX

\section{Appendix}


\subsection{Annotation}

{\flushleft\textbf{Classification Taxonomy:}} The full classification taxonomy is included in separate files, both coupled with definitions, and as a graphical visualization. 

{\flushleft\textbf{Annotation Procedure:}} Primary annotation took place over approximately 8 weeks.  Annotators were instructed to provide up to 2 labels from the full classification taxonomy (462 labels) that were appropriate for each question, and to provide the most specific label available in the taxonomy for a given question.  Of the 462 labels in the classification taxonomy, the ARC questions had non-zero counts in 406 question types. Rarely, questions were encountered by annotators that did not clearly fit into a label at the end of the taxonomy, and in these cases the annotators were instructed to choose a more generic label higher up the taxonomy that was appropriate.  This occurred when the production taxonomy failed to have specific categories for infrequent questions testing knowledge that is not a standard part of the science curriculum.  For example, the question:~\\

\textit{Which material is the best natural resource to use for making water-resistant shoes? (A) cotton (B) leather (C) plastic (D) wool}	\\

\noindent tests a student's knowledge of the water resistance of different materials.  Because this is not a standard part of the curriculum, and wasn't identified as a common topic in the training questions, the annotators tag this question as belonging to \textit{Matter $\rightarrow$ Properties of Materials}, rather than a more specific category.

Questions from the training, development, and test sets were randomly shuffled to counterbalance any learning effects during the annotation procedure, but were presented in grade order ($3^{rd}$ to $9^{th}$ grade) to reduce context switching (a given grade level tends to use a similar subset of the taxonomy -- for example, $3^{rd}$ grade questions generally do not address \textit{Chemical Equations} or \textit{Newtons $1^{st}$ Law of Motion}). 

{\flushleft\textbf{Interannotator Agreement:}} To increase quality and consistency, each annotator annotated the entire dataset of 7,787 questions.  Two annotators were used, with the lead annotator possessing previous professional domain expertise.  Annotation proceeded in a two-stage process, where in stage 1 annotators completed their annotation independently, and in stage 2 each of the questions where the annotators did not have complete agreement were manually resolved by the annotators, resulting in high-quality classification annotation. 

%
%
\begin{table}[t]
\small
\centering
\begin{tabular}{ccc}

\textbf{Classification}	&	\textbf{\# of}		& \textbf{Interannotator}\\
\textbf{Level}			&	\textbf{Classes} 	& \textbf{Agreement ($\kappa$)}\\
\hline
L1 (Coarsest)			&	9					&	0.85					\\
L2						&	88					&	0.71					\\
L3						&	243					&	0.64					\\
L4				 		&	335					&	0.60					\\
L5			 			&	379					&	0.58					\\
L6 (Finest)				&	406					&	0.58					\\

\end{tabular}
\caption{\small Interannotator Agreement at L6 (the native level the annotation was completed at), as well as agreement for truncated levels of the heirarchy from coarse to fine classification. \label{tab:interannotatoragreement}}  

\end{table}

Because each question can have up to two labels, we treat each label for a given question as a separate evaluation of interannotator agreement.  That is, for questions where both annotators labeled each question as having 1 or 2 labels, we treat this as 1 or 2 separate evaluations of interannotator agreement.  For cases where one annotator labeled as question as having 1 label, and the other annotator labeled that same question as having 2 labels, we conservatively treat this as two separate interannotator agreements where one annotator failed to specify the second label and had zero agreement on that unspecified label.

Though the classification procedure was fine-grained compared to other question classification taxonomies, containing an unusually large number of classes (406), overall raw interannotator agreement before resolution was high (Cohen's $\kappa$ = 0.58).  When labels are truncated to a maximum taxonomy depth of N, raw interannotator increases to $\kappa$ = 0.85 at the coarsest (9 class) level (see Table~\ref{tab:interannotatoragreement}). This is considered moderate to strong agreement (see McHugh \shortcite{mchugh2012interrater} for a discussion of the interpretation of the Kappa statistic).  Based on the results of an error analysis on the question classification system (see Section~\ref{sec:annotationaccuracy}), we estimate that the overall accuracy of the question classification labels after resolution is approximately 96\% . 

Annotators disagreed on 3441 (44.2\%) of questions.
Primary sources of disagreement before resolution included each annotator choosing a single category for questions requiring multiple labels (e.g. annotator 1 assigning a label of X, and annotator 2 assigning a label of Y, when the gold label was multilabel X, Y), which was observed in 18\% of disagreements. Similarly, we observed annotators choosing similar labels but at different levels of specificity in the taxonomy (e.g. annotator 1 assigning a label of \textit{Matter $\rightarrow$ Changes of State $\rightarrow$ Boiling}, where annotator 2 assigned \textit{Matter $\rightarrow$ Changes of State}), which occurred in 12\% of disagreements before resolution.



\subsection{Question Classification}

%
%
\begin{table*}[!t]
\small
\begin{center}
\begin{tabular}{llp{0.01cm}ccccccc}
	 	&	& &	\multicolumn{7}{c}{ARC Science Exams}	\\
Adapted From								&	Model								& & L1 		&	L2		&	L3		&	L4		&	L5		&	L6		& Gain (L6)	\\
\hline
											& Unigram Model 						& & 82.2	& 62.1		& 51.8		& 44.2		& 40.8		& 39.6		& \\
Li and Roth \shortcite{li2002learning} 		& Uni+Bi+POS+Hier \textit{(UBPH)}		& & 84.2	& 67.6		& 56.6		& 49.4		& 46.5		& 44.5		& Baseline\\
Van-tu et al. \shortcite{van2016improving}	& UBPH+WordNet Expansion				& & 84.1	& 67.1		& 56.4		& 49.3		& 46.4		& 44.7		& +0.2\\
Roberts et al. \shortcite{roberts2014automatically}	& UBPH+Dependencies				& & 84.7	& 68.0		& 56.5		& 49.2		& 45.6		& 44.8		& +0.3\\

He et al. \shortcite{he2015multi}			& MP-CNN								& & 84.8	& 66.3		& 56.3		& 50.7		& 46.6		& 43.5		& --\\

Khashabi et al. \shortcite{khashabi2017learning}& UBPH+Essential Terms				& & 85.9	& 69.4		& 58.7		& 51.9		& 48.4		& 48.0		& +3.5$^{*}$\\
This Work							 		& BERT-QC 								& & \textbf{90.2}	& \textbf{78.2}& \textbf{67.6}		& \textbf{60.6}	& \textbf{58.9}		& \textbf{57.8}		& \textbf{+13.5$^{*}$}\\
%
~\\
\multicolumn{2}{l}{\textit{Number of Categories}}						& &	\textit{9}	&	\textit{88}	& \textit{243} &	\textit{335}	&	\textit{379}	&	\textit{406}		\\
 
\end{tabular}
\caption{\small Performance of each question classification model, expressed in Precision@1 (P@1). * signifies a given model is significantly different from the baseline model ($p<0.01$). \label{tab:qcresultsat1}}
\end{center}

\end{table*}

\subsubsection{Precision@1}
\noindent Because of space limitations the question classification results are reported in Table~\ref{tab:results} only using Mean Average Precision (MAP).  We also include Precision@1 (P@1), the overall accuracy of the highest-ranked prediction for each question classification model, in Table~\ref{tab:qcresultsat1}.

\subsubsection{Negative Results}
{\flushleft\textbf{CNN:}} We implemented the CNN sentence classifier of Kim \shortcite{kim2014cnn}, which demonstrated near state-of-the-art performance on a number of sentence classification tasks (including TREC question classification) by using pre-trained word embeddings \cite{mikolov2013distributed} as feature extractors in a CNN model.  We adapted the original CNN non-static model to multi-label classification by changing the fully connected softmax layer to sigmoid layer to produce a sigmoid output for each label simultaneously. We followed the same parameter settings reported by Kim et al. except the learning rate, which was tuned based on the development set.  Pilot experiments did not show a performance improvement over the baseline model.

{\flushleft\textbf{Label Definitions:}}
Question terms can be mapped to categories using manual heuristics \cite[e.g.]{silva2011symbolic}.  
To mitigate sparsity and limit heuristic use, here we generated a feature comparing the cosine similarity of composite embedding vectors \cite[e.g.]{jansen14} representing question text and category definition text, using pretrained GloVe embeddings \cite{pennington2014glove}.  Pilot experiments showed that performance did not significantly improve. 

{\flushleft\textbf{Question Expansion with Hypernyms (Probase Version):}} 
One of the challenges of hypernym expansion \cite[e.g.]{huang2008question,silva2011symbolic,roberts2014automatically} is determining a heuristic for the termination depth of hypernym expansion, as in Van-tu et al. \shortcite{van2016improving}.  Because science exam questions are often grounded in specific examples (e.g. a car rolling down a hill coming to a stop due to friction), we hypothesized that knowing certain categories of entities can be important for identifying specific question types -- for example, observing that a question contains a kind of \textit{animal} may be suggestive of a \textit{Life Science} question, where similarly \textit{vehicles} or \textit{materials} present in questions may suggest questions about \textit{Forces} or \textit{Matter}, respectively.  The challenge with WordNet is that key hypernyms can be at very different depths from query terms -- for example, \textit{``cat''} is distance 10 away from \texttt{living thing}, \textit{``car''} is distance 4 away from \texttt{vehicle}, and \textit{``copper''} is distance 2 away from \texttt{material}.  Choosing a static threshold (or decaying threshold, as in Van-tu et al. \shortcite{van2016improving}) will inheriently reduce recall and limit the utility of this method of query expansion. 

To address this, we piloted a hypernym expansion experiment using the Probase taxonomy \cite{wu2012probase}, a collection of 20.7M \texttt{is-a} pairs mined from the web, in place of WordNet.  Because the taxonomic pairs in Probase come from use in naturalistic settings, links tend to jump levels in the WordNet taxonomy and be expressed in common forms.  For example, $cat \rightarrow animal$, $car \rightarrow vehicle$, and $copper \rightarrow material$, are each distance 1 in the Probase taxonomy, and high-frequency (i.e. high-confidence) taxonomic pairs. 

Similar to query expansion using WordNet Hypernyms, our pilot experiments did not observe a benefit to using Probase hypernyms over the baseline model.  An error analysis suggested that the large number of noisy and out-of-context links present in Probase may have reduced performance, and in response we constructed a filtered list of 710 key hypernym categories manually filtered from a list of hypernyms seeded using high-frequency words from an in-house corpus of 250 in-domain science textbooks.  We also did not observe a benefit to question classification over the baseline model when expanding only to this manually curated list of key hypernyms. 

\subsubsection{Additional Positive Results}

{\flushleft\textbf{Topic words:}} 
We made use of the 77 TREC word lists of Li and Roth \shortcite{li2002learning}, containing a total of 3,257 terms, as well as an in-house set of 144 word lists on general and elementary science topics mined from the web, such as \textit{ANIMALS}, \textit{VEGETABLES}, and \textit{VEHICLES}, containing a total of 29,059 words.  To mitigate sparsity, features take the form of counts for a specific topic --  detecting the words \textit{turtle} and \textit{giraffe} in a question would provide a count of 2 for the \textit{ANIMAL} feature. This provides a light form of domain-specific entity and action (e.g. types of \textit{changes}) recognition. Pilot experiments showed that this wordlist feature did add a modest performance benefit of approximately 2\% to question classification accuracy.  Taken together with our results on hypernym expansion, this suggests that manually curated wordlists can show modest benefits for question classification performance, but at the expense of substantial effort in authoring or collecting these extensive wordlists. 


\subsubsection{Additional BERT-QC Model Details}

{\flushleft\textbf{Hyperparameters:}} For each layer of the class label hierarchy, we tune the hyperparameters based on the development set.  We use the pretrained BERT-Base (uncased) checkpoint.  We use the following hyperparameters: maximum sequence length = 256, batch size = 16, learning rates: 2e-5 (L1), 5e-5 (L2-L6), epochs: 5 (L1), 25 (L2-L6).

{\flushleft\textbf{Statistics:}} We use non-parametric bootstrap resampling to compare the baseline (Li and Roth \shortcite{li2002learning} model) to all experimental models to determine significance, using 10,000 bootstrap resamples.

\subsection{Question Answering with QC Labels}

{\flushleft\textbf{Hyperparameters:}} Pilot experiments on both pre-trained BERT-Base and BERT-Large checkpoints showed similar performance benefits at the finest levels of question classification granularity (L6), but the BERT-Large model demonstrated higher overall baseline performance, and larger incremental benefits at lower levels of QC granularity, so we evaluated using that model.  We lightly tuned hyperparameters on the development set surrounding those reported by Devlin et al. \shortcite{devlin2018bert}, and ultimately settled on parameters similar to their original work, tempered by technical limitations in running the BERT-Large model on available hardware: maximum sequence length = 128, batch size = 16, learning rate: 1e-5. We report performance as the average of 10 runs for each datapoint.  The number of epochs were tuned on each run on the development set (to a maximum of 8 epochs), where most models converged to maximum performance within 5 epochs. 

{\flushleft\textbf{Preference for uncorrelated errors in multiple choice question classification:}} We primarily report QA performance using BERT-QC trained using text from only the multiple choice questions and not their answer candidates.  While this model achieved lower overall QC performance compared to the model trained with both question and multiple choice answer candidate text, it achieved slightly higher performance in the QA+QC setting.  Our error analysis in Section~\ref{sec:qcerroranalysis} shows that though models trained on both question and answer text can achieve higher QC performance, when they make QC errors, the errors tend to be highly correlated with an incorrect answer candidate, which can substantially reduce QA performance.  This is an important result for question classification in the context of multiple choice exams.
In the context of multiple choice exams, correlated noise can substantially reduce QA performance, meaning the kinds of errors that a model makes are important, and evaluating QC performance in context with QA models that make use of those QC systems is critical.  

Related to this result, we provide an analysis of the noise sensitivity of the QA+QC model for different levels of question classification prediction accuracy.  Here, we perturb gold question labels by randomly selecting a proportion of questions (between 5\% and 40\%) and randomly assigning that question a different label.  Figure \ref{fig:qaperformance1} shows that this uncorrelated noise provides roughly linear decreases in performance, and still shows moderate gains at 60\% accuracy (40\% noise) with uncorrelated noise.  This suggests that when making errors, making random errors (that are not correlated to incorrect multiple choice answers) is preferential. 

%
%
\begin{figure}[t]
	\centering
	\includegraphics[scale=0.38]{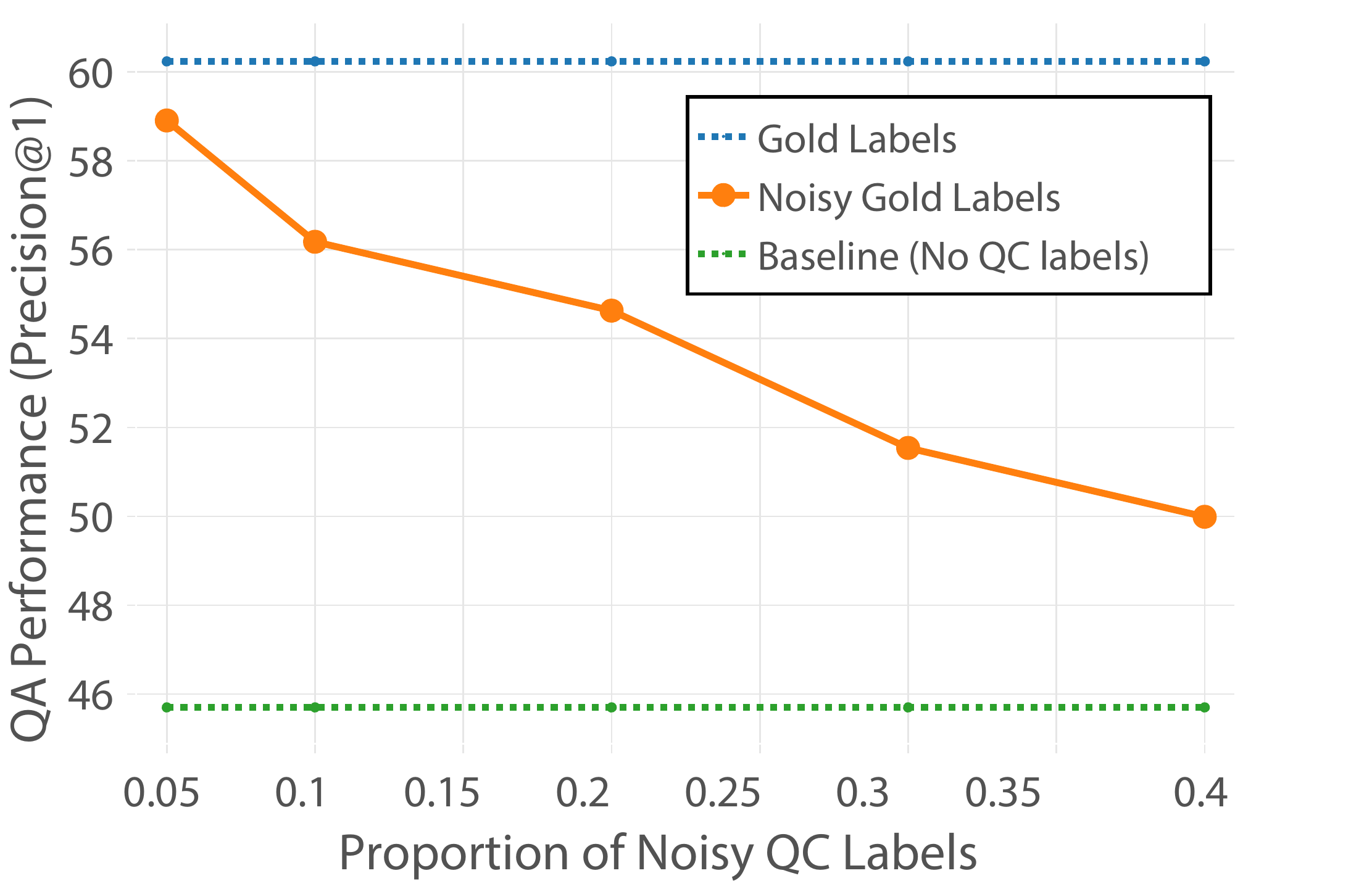}
	\caption{Analysis of noisy question classification labels on overall QA performance.  Here, the X axis represents the proportion of gold QA labels that have been randomly switched to another of the 406 possible labels at the finest level of granularity in the classification taxonomy (L6).  QA performance decreases approximately linearly as the proportion of noisy QC labels increases.  Each point represents the average of 20 experimental runs, with different questions and random labels for each run.  QA performance reported is on the development set.  Note that due to the runtime associated with this analysis, the results reported are using the BERT-Base model. \label{fig:qaperformance1} }
\end{figure}

{\flushleft\textbf{Training with predicted labels:}} We observed small gains when training the BERT-QA model with predicted QC labels.  We generate predicted labels for the training set using 5-fold crossvalidation over only the training questions. 

{\flushleft\textbf{Statistics:}} We use non-parametric bootstrap resampling to compare baseline (no label) and experimental (QC labeled) runs of the QA+QC experiment.  Because the BERT-QA model produces different performance values across successive runs, we perform 10 runs of each condition.  We then compute pairwise p-values for each of the 10 no label and QC labeled runs (generating 100 comparisons), then use Fisher's method to combine these into a final statistic.

\subsubsection{Interpretation of non-linear question answering gains between levels}

\noindent Question classification paired with question answering shows statistically significant gains of +1.7\% P@1 at L6 using predicted labels, and a ceiling gain of up to +10\% P@1 using gold labels.  The QA performance graph in Figure~\ref{fig:qaperformance} contains two deviations from the expectation of linear gains with increasing specificity, at L1 and L3.  \textit{Region at $L2 \rightarrow L3:$} On gold labels, L3 provides small gains over L2, where as L4 provides large gains over L3.  We hypothesize that this is because approximately 57\% of question labels belong to the \textit{Earth Science} or \textit{Life Science} categories which have much more depth than breadth in the standardized science curriculum, and as such these categories are primarily differentiated from broad topics into detailed problem types at levels L4 through L6. Most other curriculum categories have more breadth than depth, and show strong (but not necessarily full) differentiation at L2. \textit{Region at $L1:$} Predicted performance at L1 is higher than gold performance at L1.  We hypothesize this is because we train using predicted rather than gold labels, which provides a boost in performance. Training on gold labels and testing on predicted labels substantially reduces the difference between gold and predicted performance.

\subsubsection{Overall annotation accuracy}
\label{sec:annotationaccuracy}
\noindent Though initial raw interannotator agreement was measured at $kappa = 0.58$, to maximize the quality of the annotation the annotators performed a second pass where all disagreements were manually resolved.  
Table~\ref{tab:qcresultsat1} shows question classification performance of the BERT-QC model at 57.8\% P@1, meaning 42.2\% of the predicted labels were different than the gold labels.  The question classification error analysis in Table~\ref{tab:qcerrors1} found that of these 42.2\% of errorful predictions, 10\% of errors (4.2\% of total labels) were caused by the gold labels being incorrect.  This allows us to estimate that the overall quality of the annotation (the proportion of questions that have a correct human authored label) is approximately 96\%.

\fi

\section{Bibliographical References}
\label{main:ref}

\bibliographystyle{lrec}

\bibliography{refs}

\begin{thebibliography}{}

\bibitem[\protect\citename{Chang and Lin}2011]{chang2011libsvm}
Chang, C.-C. and Lin, C.-J.
\newblock (2011).
\newblock Libsvm: a library for support vector machines.
\newblock {\em ACM transactions on intelligent systems and technology (TIST)},
  2(3):27.

\bibitem[\protect\citename{Clark \bgroup et al.\egroup }2013]{clark:2013}
Clark, P., Harrison, P., and Balasubramanian, N.
\newblock (2013).
\newblock A study of the knowledge base requirements for passing an elementary
  science test.
\newblock In {\em Proceedings of the 2013 Workshop on Automated Knowledge Base
  Construction}, AKBC'13, pages 37--42.

\bibitem[\protect\citename{Clark \bgroup et al.\egroup
  }2016]{Clark2016CombiningRS}
Clark, P., Etzioni, O., Khot, T., Sabharwal, A., Tafjord, O., Turney, P.~D.,
  and Khashabi, D.
\newblock (2016).
\newblock Combining retrieval, statistics, and inference to answer elementary
  science questions.
\newblock In {\em Proceedings of the Thirtieth {AAAI} Conference on Artificial
  Intelligence, February 12-17, 2016, Phoenix, Arizona, {USA.}}, pages
  2580--2586.

\bibitem[\protect\citename{Clark \bgroup et al.\egroup }2018]{clark2018think}
Clark, P., Cowhey, I., Etzioni, O., Khot, T., Sabharwal, A., Schoenick, C., and
  Tafjord, O.
\newblock (2018).
\newblock Think you have solved question answering? try arc, the ai2 reasoning
  challenge.
\newblock {\em arXiv preprint arXiv:1803.05457}.

\bibitem[\protect\citename{Clark}2015]{clark:2015}
Clark, P.
\newblock (2015).
\newblock Elementary school science and math tests as a driver for {AI:} take
  the aristo challenge!
\newblock In Blai Bonet et~al., editors, {\em Proceedings of the Twenty-Ninth
  {AAAI} Conference on Artificial Intelligence, January 25-30, 2015, Austin,
  Texas, {USA.}}, pages 4019--4021. {AAAI} Press.

\bibitem[\protect\citename{De~Marneffe and Manning}2008]{de2008stanford}
De~Marneffe, M.-C. and Manning, C.~D.
\newblock (2008).
\newblock The stanford typed dependencies representation.
\newblock In {\em Coling 2008: proceedings of the workshop on cross-framework
  and cross-domain parser evaluation}, pages 1--8. Association for
  Computational Linguistics.

\bibitem[\protect\citename{Devlin \bgroup et al.\egroup }2018]{devlin2018bert}
Devlin, J., Chang, M.-W., Lee, K., and Toutanova, K.
\newblock (2018).
\newblock Bert: Pre-training of deep bidirectional transformers for language
  understanding.
\newblock {\em arXiv preprint arXiv:1810.04805}.

\bibitem[\protect\citename{Fellbaum}1998]{fellbaum1998wordnet}
Fellbaum, C.
\newblock (1998).
\newblock {\em WordNet}.
\newblock Wiley Online Library.

\bibitem[\protect\citename{Godea and Nielsen}2018]{godea2018annotating}
Godea, A. and Nielsen, R.
\newblock (2018).
\newblock Annotating educational questions for student response analysis.
\newblock In {\em Proceedings of the Eleventh International Conference on
  Language Resources and Evaluation (LREC-2018)}.

\bibitem[\protect\citename{He \bgroup et al.\egroup }2015]{he2015multi}
He, H., Gimpel, K., and Lin, J.
\newblock (2015).
\newblock Multi-perspective sentence similarity modeling with convolutional
  neural networks.
\newblock In {\em Proceedings of the 2015 Conference on Empirical Methods in
  Natural Language Processing}, pages 1576--1586.

\bibitem[\protect\citename{Hovy \bgroup et al.\egroup }2001]{hovy2001toward}
Hovy, E., Gerber, L., Hermjakob, U., Lin, C.-Y., and Ravichandran, D.
\newblock (2001).
\newblock Toward semantics-based answer pinpointing.
\newblock In {\em Proceedings of the first international conference on Human
  language technology research}, pages 1--7. Association for Computational
  Linguistics.

\bibitem[\protect\citename{Huang \bgroup et al.\egroup
  }2008]{huang2008question}
Huang, Z., Thint, M., and Qin, Z.
\newblock (2008).
\newblock Question classification using head words and their hypernyms.
\newblock In {\em Proceedings of the Conference on Empirical Methods in Natural
  Language Processing}, pages 927--936. Association for Computational
  Linguistics.

\bibitem[\protect\citename{Jansen \bgroup et al.\egroup }2014]{jansen14}
Jansen, P., Surdeanu, M., and Clark, P.
\newblock (2014).
\newblock Discourse complements lexical semantics for non-factoid answer
  reranking.
\newblock In {\em Proceedings of the 52nd Annual Meeting of the Association for
  Computational Linguistics (ACL)}.

\bibitem[\protect\citename{Jansen \bgroup et al.\egroup
  }2016]{jansen2016:COLING}
Jansen, P., Balasubramanian, N., Surdeanu, M., and Clark, P.
\newblock (2016).
\newblock What's in an explanation? characterizing knowledge and inference
  requirements for elementary science exams.
\newblock In {\em Proceedings of COLING 2016, the 26th International Conference
  on Computational Linguistics: Technical Papers}, pages 2956--2965, Osaka,
  Japan, December.

\bibitem[\protect\citename{Jansen \bgroup et al.\egroup
  }2017]{jansen2017framing}
Jansen, P., Sharp, R., Surdeanu, M., and Clark, P.
\newblock (2017).
\newblock Framing qa as building and ranking intersentence answer
  justifications.
\newblock {\em Computational Linguistics}.

\bibitem[\protect\citename{Jansen \bgroup et al.\egroup
  }2018]{jansen2018worldtree}
Jansen, P., Wainwright, E., Marmorstein, S., and Morrison, C.
\newblock (2018).
\newblock Worldtree: A corpus of explanation graphs for elementary science
  questions supporting multi-hop inference.
\newblock In {\em Proceedings of the Eleventh International Conference on
  Language Resources and Evaluation (LREC-2018)}.

\bibitem[\protect\citename{Khashabi \bgroup et al.\egroup
  }2016]{Khashabi:2016TableILP}
Khashabi, D., Khot, T., Sabharwal, A., Clark, P., Etzioni, O., and Roth, D.
\newblock (2016).
\newblock Question answering via integer programming over semi-structured
  knowledge.
\newblock In {\em Proceedings of the International Joint Conference on
  Artificial Intelligence}, IJCAI'16, pages 1145--1152.

\bibitem[\protect\citename{Khashabi \bgroup et al.\egroup
  }2017]{khashabi2017learning}
Khashabi, D., Khot, T., Sabharwal, A., and Roth, D.
\newblock (2017).
\newblock Learning what is essential in questions.
\newblock In {\em Proceedings of the 21st Conference on Computational Natural
  Language Learning (CoNLL 2017)}, pages 80--89.

\bibitem[\protect\citename{Khot \bgroup et al.\egroup
  }2015]{Khot2015ExploringML}
Khot, T., Balasubramanian, N., Gribkoff, E., Sabharwal, A., Clark, P., and
  Etzioni, O.
\newblock (2015).
\newblock Exploring markov logic networks for question answering.
\newblock In {\em EMNLP}.

\bibitem[\protect\citename{Khot \bgroup et al.\egroup }2017]{Khot:ACL2017}
Khot, T., Sabharwal, A., and Clark, P.
\newblock (2017).
\newblock Answering complex questions using open information extraction.
\newblock In {\em Proceedings of the 55th Annual Meeting of the Association for
  Computational Linguistics, {ACL} 2017, Vancouver, Canada, July 30 - August 4,
  Volume 2: Short Papers}, pages 311--316.

\bibitem[\protect\citename{Kim}2014]{kim2014cnn}
Kim, Y.
\newblock (2014).
\newblock Convolutional neural networks for sentence classification.
\newblock In {\em Proceedings of the 2014 conference on empirical methods in
  natural language processing (EMNLP)}, pages 1746--1751.

\bibitem[\protect\citename{Lally \bgroup et al.\egroup
  }2012]{lally2012question}
Lally, A., Prager, J.~M., McCord, M.~C., Boguraev, B.~K., Patwardhan, S., Fan,
  J., Fodor, P., and Chu-Carroll, J.
\newblock (2012).
\newblock Question analysis: How watson reads a clue.
\newblock {\em IBM Journal of Research and Development}, 56(3.4):2--1.

\bibitem[\protect\citename{Lei \bgroup et al.\egroup }2018]{lei2018novel}
Lei, T., Shi, Z., Liu, D., Yang, L., and Zhu, F.
\newblock (2018).
\newblock A novel cnn-based method for question classification in intelligent
  question answering.
\newblock In {\em Proceedings of the 2018 International Conference on
  Algorithms, Computing and Artificial Intelligence}, page~54. ACM.

\bibitem[\protect\citename{Lesk}1986]{lesk1986automatic}
Lesk, M.
\newblock (1986).
\newblock Automatic sense disambiguation using machine readable dictionaries:
  how to tell a pine cone from an ice cream cone.
\newblock In {\em Proceedings of the 5th annual international conference on
  Systems documentation}, pages 24--26. ACM.

\bibitem[\protect\citename{Li and Roth}2002]{li2002learning}
Li, X. and Roth, D.
\newblock (2002).
\newblock Learning question classifiers.
\newblock In {\em Proceedings of the 19th international conference on
  Computational linguistics-Volume 1}, pages 1--7. Association for
  Computational Linguistics.

\bibitem[\protect\citename{Li and Roth}2006]{li2006learning}
Li, X. and Roth, D.
\newblock (2006).
\newblock Learning question classifiers: the role of semantic information.
\newblock {\em Natural Language Engineering}, 12(3):229--249.

\bibitem[\protect\citename{Liu \bgroup et al.\egroup }2011]{liu2011toward}
Liu, F., Antieau, L.~D., and Yu, H.
\newblock (2011).
\newblock Toward automated consumer question answering: Automatically
  separating consumer questions from professional questions in the healthcare
  domain.
\newblock {\em Journal of biomedical informatics}, 44(6):1032--1038.

\bibitem[\protect\citename{Madabushi and Lee}2016]{madabushi2016high}
Madabushi, H.~T. and Lee, M.
\newblock (2016).
\newblock High accuracy rule-based question classification using question
  syntax and semantics.
\newblock In {\em COLING}, pages 1220--1230.

\bibitem[\protect\citename{Manning \bgroup et al.\egroup }2008]{manning08}
Manning, C.~D., Raghavan, P., and Sch\"{u}tze, H.
\newblock (2008).
\newblock {\em Introduction to Information Retrieval}.
\newblock Cambridge University Press.

\bibitem[\protect\citename{McHugh}2012]{mchugh2012interrater}
McHugh, M.~L.
\newblock (2012).
\newblock Interrater reliability: the kappa statistic.
\newblock {\em Biochemia medica: Biochemia medica}, 22(3):276--282.

\bibitem[\protect\citename{Mikolov \bgroup et al.\egroup
  }2013]{mikolov2013distributed}
Mikolov, T., Sutskever, I., Chen, K., Corrado, G.~S., and Dean, J.
\newblock (2013).
\newblock Distributed representations of words and phrases and their
  compositionality.
\newblock In {\em Advances in neural information processing systems}, pages
  3111--3119.

\bibitem[\protect\citename{Minsky}1986]{Minsky:1986:SM}
Minsky, M.
\newblock (1986).
\newblock {\em The Society of Mind}.
\newblock Simon \& Schuster, Inc., New York, NY, USA.

\bibitem[\protect\citename{Mishra \bgroup et al.\egroup
  }2013]{mishra2013question}
Mishra, M., Mishra, V.~K., and Sharma, H.
\newblock (2013).
\newblock Question classification using semantic, syntactic and lexical
  features.
\newblock {\em International Journal of Web \& Semantic Technology}, 4(3):39.

\bibitem[\protect\citename{Moldovan \bgroup et al.\egroup
  }2003]{Moldovan:2003:PIE:763693.763694}
Moldovan, D., Pa\c{s}ca, M., Harabagiu, S., and Surdeanu, M.
\newblock (2003).
\newblock Performance issues and error analysis in an open-domain question
  answering system.
\newblock {\em ACM Trans. Inf. Syst.}, 21(2):133--154, April.

\bibitem[\protect\citename{Neves and Kraus}2016]{neves2016biomedlat}
Neves, M. and Kraus, M.
\newblock (2016).
\newblock Biomedlat corpus: Annotation of the lexical answer type for
  biomedical questions.
\newblock {\em OKBQA 2016}, page~49.

\bibitem[\protect\citename{Nguyen \bgroup et al.\egroup }2016]{nguyen2016ms}
Nguyen, T., Rosenberg, M., Song, X., Gao, J., Tiwary, S., Majumder, R., and
  Deng, L.
\newblock (2016).
\newblock Ms marco: A human generated machine reading comprehension dataset.
\newblock {\em arXiv preprint arXiv:1611.09268}.

\bibitem[\protect\citename{Pan \bgroup et al.\egroup }2019]{pan2019improving}
Pan, X., Sun, K., Yu, D., Ji, H., and Yu, D.
\newblock (2019).
\newblock Improving question answering with external knowledge.
\newblock {\em arXiv preprint arXiv:1902.00993}.

\bibitem[\protect\citename{Patrick and Li}2012]{patrick2012ontology}
Patrick, J. and Li, M.
\newblock (2012).
\newblock An ontology for clinical questions about the contents of patient
  notes.
\newblock {\em Journal of Biomedical Informatics}, 45(2):292--306.

\bibitem[\protect\citename{Pennington \bgroup et al.\egroup
  }2014]{pennington2014glove}
Pennington, J., Socher, R., and Manning, C.
\newblock (2014).
\newblock Glove: Global vectors for word representation.
\newblock In {\em Proceedings of the 2014 conference on empirical methods in
  natural language processing (EMNLP)}, pages 1532--1543.

\bibitem[\protect\citename{Qiu and Frei}1993]{qiu1993concept}
Qiu, Y. and Frei, H.-P.
\newblock (1993).
\newblock Concept based query expansion.
\newblock In {\em Proceedings of the 16th annual international ACM SIGIR
  conference on Research and development in information retrieval}, pages
  160--169. ACM.

\bibitem[\protect\citename{Rajpurkar \bgroup et al.\egroup
  }2016]{rajpurkar2016squad}
Rajpurkar, P., Zhang, J., Lopyrev, K., and Liang, P.
\newblock (2016).
\newblock Squad: 100,000+ questions for machine comprehension of text.
\newblock {\em arXiv preprint arXiv:1606.05250}.

\bibitem[\protect\citename{Rao \bgroup et al.\egroup }2016]{rao2016noise}
Rao, J., He, H., and Lin, J.
\newblock (2016).
\newblock Noise-contrastive estimation for answer selection with deep neural
  networks.
\newblock In {\em Proceedings of the 25th ACM International on Conference on
  Information and Knowledge Management}, pages 1913--1916. ACM.

\bibitem[\protect\citename{Roberts \bgroup et al.\egroup
  }2014]{roberts2014automatically}
Roberts, K., Kilicoglu, H., Fiszman, M., and Demner-Fushman, D.
\newblock (2014).
\newblock Automatically classifying question types for consumer health
  questions.
\newblock In {\em AMIA Annual Symposium Proceedings}, volume 2014, page 1018.
  American Medical Informatics Association.

\bibitem[\protect\citename{Sarrouti \bgroup et al.\egroup
  }2015]{sarrouti2015biomedical}
Sarrouti, M., Lachkar, A., and Ouatik, S. E.~A.
\newblock (2015).
\newblock Biomedical question types classification using syntactic and rule
  based approach.
\newblock In {\em 2015 7th International Joint Conference on Knowledge
  Discovery, Knowledge Engineering and Knowledge Management (IC3K)}, volume~1,
  pages 265--272. IEEE.

\bibitem[\protect\citename{Silva \bgroup et al.\egroup
  }2011]{silva2011symbolic}
Silva, J., Coheur, L., Mendes, A.~C., and Wichert, A.
\newblock (2011).
\newblock From symbolic to sub-symbolic information in question classification.
\newblock {\em Artificial Intelligence Review}, 35(2):137--154.

\bibitem[\protect\citename{Tsoumakas and Katakis}2007]{tsoumakas2007multi}
Tsoumakas, G. and Katakis, I.
\newblock (2007).
\newblock Multi-label classification: An overview.
\newblock {\em International Journal of Data Warehousing and Mining (IJDWM)},
  3(3):1--13.

\bibitem[\protect\citename{Tu}2018]{tu2018experimental}
Tu, Z.
\newblock (2018).
\newblock An experimental analysis of multi-perspective convolutional neural
  networks.
\newblock {\em University of Waterloo Master's Thesis}.

\bibitem[\protect\citename{Van-Tu and Anh-Cuong}2016]{van2016improving}
Van-Tu, N. and Anh-Cuong, L.
\newblock (2016).
\newblock Improving question classification by feature extraction and
  selection.
\newblock {\em Indian Journal of Science and Technology}, 9(17).

\bibitem[\protect\citename{Voorhees and Tice}2000]{voorhees2000building}
Voorhees, E.~M. and Tice, D.~M.
\newblock (2000).
\newblock Building a question answering test collection.
\newblock In {\em Proceedings of the 23rd annual international ACM SIGIR
  conference on Research and development in information retrieval}, pages
  200--207. ACM.

\bibitem[\protect\citename{Wasim \bgroup et al.\egroup }2019]{wasim2019multi}
Wasim, M., Asim, M.~N., Khan, M. U.~G., and Mahmood, W.
\newblock (2019).
\newblock Multi-label biomedical question classification for lexical answer
  type prediction.
\newblock {\em Journal of biomedical informatics}, page 103143.

\bibitem[\protect\citename{Wu \bgroup et al.\egroup }2012]{wu2012probase}
Wu, W., Li, H., Wang, H., and Zhu, K.~Q.
\newblock (2012).
\newblock Probase: A probabilistic taxonomy for text understanding.
\newblock In {\em Proceedings of the 2012 ACM SIGMOD International Conference
  on Management of Data}, pages 481--492. ACM.

\bibitem[\protect\citename{Xia \bgroup et al.\egroup }2018]{xia2018novel}
Xia, W., Zhu, W., Liao, B., Chen, M., Cai, L., and Huang, L.
\newblock (2018).
\newblock Novel architecture for long short-term memory used in question
  classification.
\newblock {\em Neurocomputing}, 299:20--31.

\bibitem[\protect\citename{Yu and Cao}2008]{yu2008automatically}
Yu, H. and Cao, Y.-g.
\newblock (2008).
\newblock Automatically extracting information needs from ad hoc clinical
  questions.
\newblock In {\em AMIA annual symposium proceedings}, volume 2008, page~96.
  American Medical Informatics Association.

\bibitem[\protect\citename{Zellers \bgroup et al.\egroup
  }2018]{zellers2018swag}
Zellers, R., Bisk, Y., Schwartz, R., and Choi, Y.
\newblock (2018).
\newblock Swag: A large-scale adversarial dataset for grounded commonsense
  inference.
\newblock {\em arXiv preprint arXiv:1808.05326}.

\end{thebibliography}


\end{document}